\documentclass[10pt,twocolumn,letterpaper]{article}

\usepackage{iccv}
\usepackage{times}
\usepackage{epsfig}
\usepackage{graphicx}
\usepackage{amsmath}
\usepackage{amssymb}
\usepackage{amsthm}
\usepackage{ulem}

\usepackage{booktabs}
\usepackage{soul}
\usepackage{xcolor}
\usepackage{multicol}
\usepackage{multirow} 
\usepackage{xcolor}
\usepackage{colortbl}
\usepackage[ruled, linesnumbered]{algorithm2e}
\usepackage{tcolorbox} 

\definecolor{blockcolor}{RGB}{169, 181, 223} 
\definecolor{keywordcolor}{RGB}{179, 216, 168} 
\definecolor{commentcolor}{RGB}{0, 153, 0}   

\newtcbox{\myblock}[1][blockcolor]{
    colback=#1, 
    colframe=white, 
    boxrule=0.0pt, 
    arc=0pt, 
    left=0pt, right=0pt, top=0pt, bottom=0pt, 
    boxsep=0pt,
}

\theoremstyle{plain} 
\newtheorem{theorem}{Theorem}[section]
\newtheorem{proposition}[theorem]{Proposition}

\newtheorem{remark}{Remark}[section] 

\definecolor{iccvblue}{rgb}{0.21,0.49,0.74}
\usepackage[pagebackref,breaklinks,colorlinks,allcolors=iccvblue]{hyperref}

\iccvfinalcopy 

\ificcvfinal\fi

\begin{document}

\title{Taming Flow Matching with Unbalanced Optimal Transport into Fast Pansharpening}

\author{Zihan Cao\footnotemark[1]\\
UESTC\\
{\tt\small iamzihan666@gmail.com}
\and
Yu Zhong\footnotemark[1]\\
UESTC\\
{\tt\small yuuzhong1011@gmail.com}
\and
Liang-Jian Deng\footnotemark[2]\\
UESTC\\
{\tt\small liangjian.deng@uestc.edu.cn}
}

\renewcommand{\thefootnote}{\fnsymbol{footnote}}

\maketitle
\ificcvfinal\thispagestyle{empty}\fi
\footnotetext[1]{Equal Contribution.}
\footnotetext[2]{Corresponding author.}

\begin{abstract}
Pansharpening, a pivotal task in remote sensing for fusing high-resolution panchromatic and multispectral imagery, has garnered significant research interest. Recent advancements employing diffusion models based on stochastic differential equations (SDEs) have demonstrated state-of-the-art performance. However, the inherent multi-step sampling process of SDEs imposes substantial computational overhead, hindering practical deployment. While existing methods adopt efficient samplers, knowledge distillation, or retraining to reduce sampling steps (\textit{e.g.}, from 1,000 to fewer steps), such approaches often compromise fusion quality.
In this work, we propose the Optimal Transport Flow Matching (OTFM) framework, which integrates the dual formulation of unbalanced optimal transport (UOT) to achieve one-step, high-quality pansharpening. Unlike conventional OT formulations that enforce rigid distribution alignment, UOT relaxes marginal constraints to enhance modeling flexibility, accommodating the intrinsic spectral and spatial disparities in remote sensing data. Furthermore, we incorporate task-specific regularization into the UOT objective, enhancing the robustness of the flow model.
The OTFM framework enables simulation-free training and single-step inference while maintaining strict adherence to pansharpening constraints. Experimental evaluations across multiple datasets demonstrate that OTFM matches or exceeds the performance of previous regression-based models and leading diffusion-based methods while only needing \textbf{one sampling step}. Codes are available at \url{https://github.com/294coder/PAN-OTFM}.
\end{abstract}

\section{Introduction}
\begin{figure}[t]
    \centering
    \includegraphics[width=\linewidth]{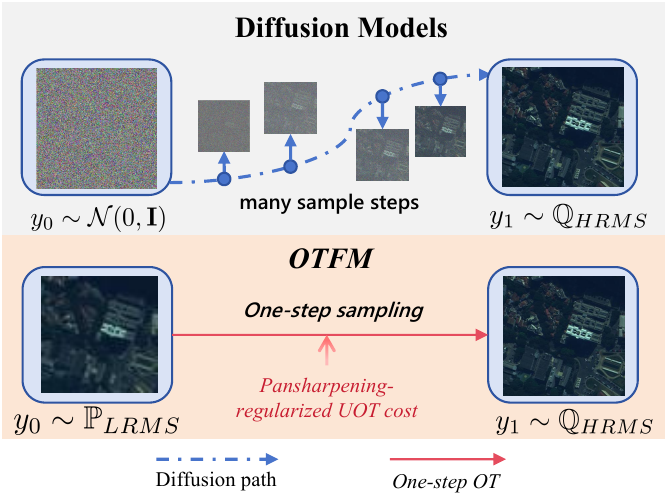}
    \caption{\textbf{Key distinction from previous diffusion models}. Traditional diffusion models typically sample from a Gaussian distribution, requiring numerous iterative steps (\textit{e.g.}, 1000) to achieve results. In contrast, our OTFM harnesses the power of unbalanced optimal transport, enabling high-quality pansharpening with just \textit{one-step} sampling step.}
    \label{fig:teaser}
    \vspace{-2em}
\end{figure}

Multispectral images, captured by collecting different wavelengths reflected by various surface objects, can reveal more comprehensive spectral and chromatic characteristics of a region, thereby reflecting more complete properties of surface features \cite{ciotola2024hyperspectral}. However, constrained by physical imaging limitations, remote sensing satellites can only acquire low-resolution multispectral images (LRMS). Pansharpening generates high-resolution multispectral images (HRMS) which are unattainable under practical imaging conditions by fusing single-channel Panchromatic Images (PAN) with high spatial resolution and LRMS. This technique has garnered widespread attention from both the research community and commercial companies. Most satellites, such as World-View3, GaoFen-2, and QuickBird, are capable of simultaneously capturing PAN and LRMS data.

Existing pansharpening methods can be categorized into two groups: traditional methods (\textit{e.g.,} component substitution methods) \cite{meng2016pansharpening, vivone2017regression,wu2023lrtcfpan} and deep learning approaches \cite{meng2023pandiff, deng2019fusion, jin2022lagconv}.
The first three conventional methods are often constrained by technical limitations, manifesting spectral and spatial distortion issues. Due to the development of deep learning, lots of deep models are proposed in the aspects of regression-based \cite{deng2019fusion,jin2022lagconv,HFIN}, unfolding-based \cite{pancsc,memoryAugPan}, and generative-based models \cite{ssdiff,cao2023ddif,meng2023pandiff}. 
Regression-based methods take LRMS and PAN as inputs and utilize residual learning to capture high-frequency information, achieving better performance than traditional methods. However, these approaches often suffer from poor interpretability and generalization. Unfolding methods leverage the concept of iterative optimization by unfolding the model, which can enhance model performance and provide interpretability. Nevertheless, unfolding with a naive for-loop cannot theoretically guarantee convergence approximation, and the unfolded model significantly expands the computational graph, leading to a substantial increase in inference overhead. Methods based on generative models, for example, GANs \cite{ma2020pan}, employ a discriminator to distinguish between degraded and real PAN and LRMS, thereby improving the generator's output quality and offering sufficient flexibility. However, these methods often encounter training instability and are prone to mode collapse.

Recent advances in generative modeling have witnessed diffusion models \cite{ho2020denoising,nichol2021improved,rombach2022high} emerge as powerful tools, with pioneering work \cite{meng2023pandiff,ssdiff} successfully adapting these architectures to pansharpening tasks. These implementations typically employ LRMS and PAN images as conditional inputs to guide the diffusion process. Subsequent innovations like Flow Matching (FM) \cite{lipman2022flow} and Rectify Flow \cite{liu2022flow} have streamlined the training paradigm through linear interpolation-based time scheduling, establishing theoretical equivalence to conventional diffusion models when initialized with Gaussian distributions.
Nevertheless, both diffusion-based and FM-based approaches rely on stochastic or ordinary differential equations (SDEs/ODEs) for image synthesis, requiring computationally intensive sampling processes with numerous number of network evaluations (NFEs). While acceleration strategies including advanced samplers \cite{karras2022elucidating,shaul2023bespoke} and knowledge distillation techniques \cite{liu2022flow,meng2023distillation} have been developed to reduce NFEs, they face critical limitations: 1) either necessitate complete network retraining or 2) incur inevitable performance compromises.

In the spirit of directly reducing NFEs during training, we leverage the theory of unbalanced optimal transport (UOT) to derive its dual formulation, and further parameterizing it into a mapping network and a potential network. To further constrain the marginals of UOT, we formulate the pansharpening task as an integration of the UOT cost into the UOT framework, enhancing generalization performance. Subsequently, we adapt the UOT loss into a training diagram suitable for flow matching (FM), stabilizing the training process while achieving \textit{one-step} generation with FM, eliminating the need for post-training or designing complex samplers. To summarize, our contributions are as follows:
\begin{itemize}
    \item We derive a dual formulation of UOT specifically tailored for the pansharpening task, which facilitates efficient one-step sampling;
    \item We integrate a regularization term customized for the pansharpening task into the UOT cost within the flow matching framework. This approach enhances both the stability of the training process and the generalization capability of the model;
    \item Our training framework, termed OTFM, enables the sampling of HRMS in \textit{a single step} while achieving performance that matches or surpasses that of previous diffusion-based methods.
\end{itemize}

\section{Preliminary}
We present the preliminaries of pansharpening and diffusion models in Suppl. Sect. {\color{iccvblue} 1}, and review the basic concepts of optimal transport and flow matching in this section.

\subsection{Optimal Transport}
\newcommand{\bbP}{\mathbb P}
\newcommand{\bbQ}{\mathbb Q}

In this section, we introduce the background of optimal transport, including the general formulations of the Monge \cite{monge1781memoire} and Kantorovich OT problems \cite{kantorovitch1958translocation}. We begin by briefly defining the notation.
There are two distributions $\mathbb P \in \mathcal P(Y)$ and $\mathbb Q \in \mathcal P(X)$. $\mathcal P(X)$ and $\mathcal P(Y)$ represent the set of distributions on the compact metric space. Let $T$ be a push-forward operator that works to push a probability mass to another, \textit{i.e.}, $T_{\#}\mathbb P=\mathbb Q$. $\Pi(\mathbb P, \mathbb Q)$ denotes the set of joint probability
distributions on $X\times Y$ whose marginals are $\mathbb P$ and $\mathbb Q$, respectively. Let $\pi \in \mathcal{M}_{+}(X \times Y)$ denote the set of joint positive measures, with $\pi_0(x)$ and $\pi_1(y)$ representing the marginals with respect to $X$ and $Y$, respectively. A cost function $c(x, y)$ is defined on $X\times Y$ and $x\in X,y\in Y$. In this paper, we consider $X=Y\subset \mathbb R^d$ with a quadratic cost $c(x,y)= \|x-y\|_2^2$, where $d$ is the dimension of the data.

Traditional OT formulations have two formats, including Monge and Kantorovich formulations.
The Monge formulation is defined as,
\begin{equation}
    C_{Monge}(\bbP, \bbQ)=\inf_{T_{\#}\bbP =\bbQ}\int_Y c(x,T(x))d\bbP(x). \label{eq: monge-ot}
\end{equation}
The optimal map $T^*$ of Monge formulation takes over all transport maps $T: Y\to X$. The constraint $T_{\#}\bbP =\bbQ$ is somehow too strict, then can be relaxed into Kantorovich formulation,
\begin{equation}
    C_{Kant}(\bbP, \bbQ)=\inf_{\pi\in\Pi (\bbP, \bbQ)}\int_{X\times Y}c(x,y)d\pi(x,y).
    \label{eq: kant-ot}
\end{equation}
The optimal coupling $\pi^*$ is taken over all transport plans on $X\times Y$ whose marginals are $\bbP$ and $\bbQ$.

Recently, many works have introduced OT into image restoration or translation tasks. For instance, \cite{korotin2022neural} formulated the Kantorovich formulation as a minimax problem for image translation; \cite{gu2023optimal} computed OT maps within a dual framework for image super-resolution; \cite{wang2022optimal} relaxed the Monge formulation into a Wasserstein-1 penalty to constrain the push-forward operator for image denoising; and LightSB \cite{korotin2023light} introduced entropy-regularized OT, linking it to the Schr\"odinger bridge \cite{leonard2013survey} for fast bridge matching. However, these methods are often constrained by the OT formulations, leading to issues such as reduced network expressiveness \cite{amos2017input} and unstable training \cite{korotin2022neural,asadulaev2022neural}.

\subsection{Flow Matching}
Flow matching \cite{liu2022flow,lipman2022flow}, as a generative method, has proven highly effective in producing high-quality images \cite{lipman2022flow,dao2023flow}, videos \cite{jin2024pyramidal}, text \cite{gat2025discrete}, and molecular structures \cite{dunn2024mixed}.
Compared to diffusion models \cite{ho2020denoising,nichol2021improved}, flow matching offers a simpler implementation of the training and sampling processes. Specifically, during training, we sample $y_0,y_1$ from prior and data distributions $\bbP,\bbQ$, and then sample $t \in \mathcal{U}(0,1)$ to construct $y_t$ through a flow path interpolation,
\begin{equation}
y_t = t y_0 + (1-t)y_1,
\end{equation}
where $y_0$ is sampled from the standard Gaussian distribution. 
We can train a velocity network $s_{\theta,t}(y_t,t)$ using the following flow loss:
\begin{equation}
\mathcal{L}_{\text{flow}} = \mathbb E_{y_0\in \bbP, y_1\in \bbQ,t\in \mathcal N(0,1) }\|s_{\theta,t}(y_t,t) - (y_1-y_0)\|_2^2,
\label{eq: flow-matching-loss}
\end{equation}
where $\bbP$ and $\bbQ$ are two interpolated distributions.
After the network converges, we can efficiently perform ODE sampling using an ODE sampler (\textit{e.g.}, Euler sampler):
\begin{equation}
dy_t = s_{\theta,t}(y_t,t)dt.
\end{equation}
When generating samples using flow matching with an ODE sampler, it typically requires 100 to 1000 sampling steps to achieve reasonable results. Even with the use of more efficient ODE samplers \cite{shaul2023bespoke}, this issue can only be slightly alleviated. Recently, distillation methods have proven highly effective in reducing the number of sampling steps, such as online distillation \cite{song2023consistency,berthelot2023tract}, offline ReFlow \cite{liu2022flow,kim2024simple}. However, these methods either load multiple teacher/student networks to optimize or require offline generation of a large number of (noise, generated image) data pairs. Therefore, a key question arises: \textit{can we directly obtain a model capable of one-step sampling during a single training process?}

\section{Methodology}
In this section, we first introduce how to formulate pansharpening under the flow matching framework. Then, we incorporate unbalanced optimal transport (UOT) into flow matching to enable one-step generation. Finally, we present network architectures of our proposed OTFM.

\subsection{Pansharpening as Flow Matching}
For previous deep regression-based methods \cite{deng2019fusion,jin2022lagconv,dcfnet}, one can formulate the pansharpening problem as,
\begin{equation}
    y = T_\theta(m,p),
\end{equation}
where $p \in P$, $m \in M$ are PAN, LRMS and in distribution $\mathbb C$. $y \in \bbQ$ is the fused HRMS. They are coupled in a product distribution $\mathbb Q\times \mathbb C$. $T_\theta$ is implemented as a deep network that takes $p$ and $m$ as input.
Under the flow matching framework, the pansharpening problem is broken down into many ODE sampling steps, formulated as,
\begin{equation}
    dy_t=s_{\theta,t}(y_t,t,m,p)dt,
\end{equation}
where $y_0 \in \mathcal N(0,\mathbf I)$, $y_1 \in \bbQ$ is the HRMS distribution, and $s_\theta$ is the velocity network.
Intuitively, we can turn the network $T_\theta$ to be timestep, PAN, and LRMS conditioned to learn the flow velocity $s$ using the flow loss.

However, challenges persist, such as network training errors \cite{karras2022elucidating}, discretization errors during sampling \cite{song2020denoising}, and accumulated errors caused by varying flow schedulers \cite{sabour2024align,park2024textit} that result in divergent sampling paths. These issues create a dilemma between sampling efficiency and generation quality: \textit{achieving high-quality generation requires numerous sampling steps, while directly reducing the number of steps exacerbates errors and degrades the performance}.

\subsection{Pansharpening Regularized Unbalanced OT}
In this subsection, we introduce how to leverage the \textit{dual formulation} of UOT to construct an optimization framework and further introduce the \textit{spatial and spectral regularization} in the UOT cost function.
This framework is then integrated into the flow matching training framework, culminating in a simulation-free OTFM training and sampling paradigm that supports one-step sampling while preserving task-specific regularization.

\subsubsection{UOT Generative Modeling}
\begin{figure*}
    \centering
    \includegraphics[width=\linewidth]{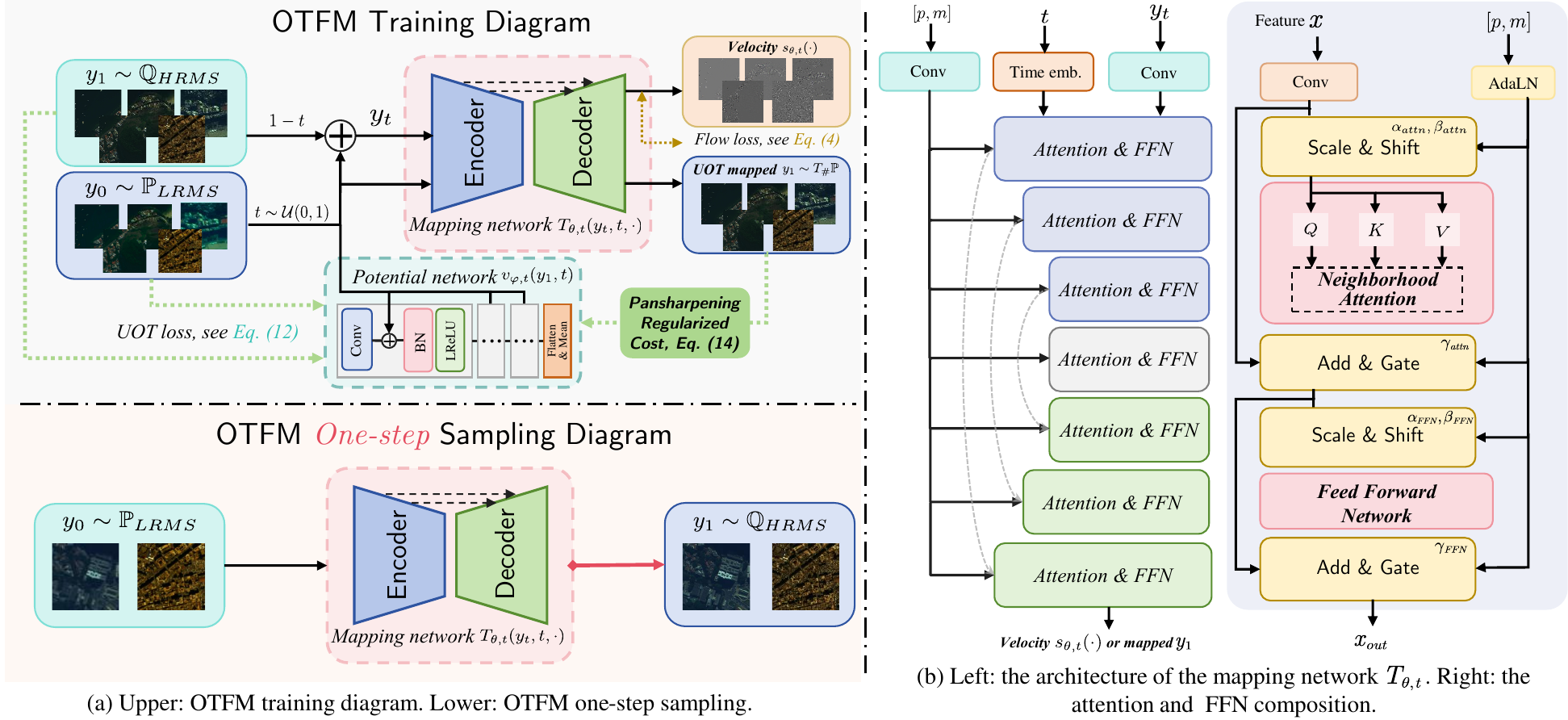}
    \caption{\textbf{(a) Training and \textit{one-step} sampling diagrams of the proposed OTFM.} Due the flow matching velocity construction (see Eq.~\eqref{eq: flow-matching-loss}), the UOT mapped $\hat y_1$ can be simply obtained from the predicted velocity $s_{\theta,t}(y_t,t,\cdot)$. A potential network $v_{\varphi}(\cdot)$ is parameterized from UOT \textit{dual} formulation (see Prop.~\ref{prop: dual-form-uot}) and supports training a one-step mapping network. Note that conditions PAN and LRMS in the mapping network are omitted for simplicity. \textbf{(b) Mapping network designs for pansharpening}. The network is adopted as a U-net~\cite{ronneberger2015u} architecture. To inject conditions (\textit{i.e.}, $[p,m]$), the zero-AdaLN is used to scale, shift, gate the feature $x$.}
    \label{fig:otfm-framework}
    \vspace{-1em}
\end{figure*}

Under some mild assumption of $(X,\bbP)$ and $(Y,\bbQ)$ \cite{villani2008optimal}, the minimizer $\pi^*$ of Kantorovich problem~\eqref{eq: kant-ot} always exists, and the dual form is given as,
\newcommand{\td}{\text{d}}
\begin{equation}
    C(\bbP, \bbQ)=\sup\left[\int_X u(x) d\bbP(x)+\int_Yv(y)d\bbQ (y)\right],
    \label{eq: ot-dual}
\end{equation}
where the $\sup$ is taken under ${\bbP(x)+\bbQ(y)\leq c(x,y)}$. $u$ and $v$ are density of distribution $\bbP$ and $\bbQ$. Assuming $u=-v$ and $u$ is 1-Lipschitz \cite{villani2008optimal}, the Eq.~\eqref{eq: ot-dual} can be reformulated using \textit{Kantorovich-Rubinstein duality} \cite{kantorovitch1958translocation} as,
\begin{equation}
    C(\bbP, \bbQ)=\sup_{v\in L^1(\bbQ)}\left[\int_X v^c(x)d\bbP(x)+\int_Y v(y)d \bbQ (y)\right],
    \label{eq: c-transformation}
\end{equation}
where $v^c$ denotes the $c$-transform \cite{backhoff2019existence} of $v$, and meets $v^c(x)=\inf_{y\in Y}(c(x,y)-v(y))$. 

To relax the hard marginal constraints of Eq.~\eqref{eq: monge-ot} which cause performance drop and optimization unstableness, UOT \cite{chizat2018unbalanced} can be introduced. Let an entropy function $f:[0,\infty)\to [0,\infty)$ be convex, non-negative, and lower semi-continuous and $D_f(\mu|\nu)$ be the $f$-divergence for the case that the measure $\mu$ is not absolutely continuous w.r.t measure $\nu$. Thus, we can relax the marginal constraint by defining the UOT formulation $ C_{UOT}(\bbP,\bbQ)$,
\begin{equation}    
\begin{aligned}
    \inf_{\pi \in \Pi(\bbP,\bbQ)}\int_{X\times Y} (x,y) d\pi(x,y)+D_f(\pi_0|\bbP)+D_f(\pi_1|\bbQ). \notag
\end{aligned}
\end{equation}
We can summarize some properties of UOT and further, according to \cite{gallouet2021regularity} and Eq.~\eqref{eq: kant-ot}, the \textit{dual form} of UOT can be proposed with the proof in Suppl. Sect. {\color{iccvblue} 2.1}.

\begin{remark}[Properties of UOT]
    1) UOT can transport any measurable mass because it relaxes the marginal constraint; 2) UOT is not sensitive to out-of-distribution (OOD) samples, which limits traditional OT methods. If a sample is OOD, UOT can adaptively move the marginals.
\end{remark}
\begin{proposition}[Dual formulation of UOT] \label{prop: dual-form-uot}
    The UOT dual formulation, $C_{UOT}(\bbP,\bbQ)$, can be obtained by using the $c$-transform:
    \begin{equation}
       \sup_{v\in C} \left[\int_X -f(-v^c(x))d\bbP+\int_Y-f(-v(y))d\bbQ\right], \label{eq: dual-form-uot}
    \end{equation}
    where $f$ is the entropy function and $v^c$ is the $c$-transformation of $v$ in Eq.~\eqref{eq: c-transformation}.
\end{proposition}

\begin{remark}[Possible choices of function $f$]
    UOT only constraints $f$ to be convex, non-decreasing, lower semi-continuous, and differentiable functions. Since the conjugate of a function is convex and lower semi-continuous, we empirically choose $f:=\exp(\cdot)$ to be a general case.
\end{remark}

At this time, we can use the dual form to construct a generative framework. Then we introduce the mapping network $T_\theta$ to approximate the $v^c$ as following,
\begin{equation}
    T_{\theta}(x)=\mathop{\arg\inf}_{y\in Y}[c(x,T_\theta(x))-v(T_\theta(x))].
    \label{eq: T-objective}
\end{equation}
For $v$, the UOT dual-formulation objective $\mathcal K(v)$ can be derived from the supremum of Eq.~\eqref{eq: dual-form-uot} and Eq.~\eqref{eq: T-objective},
\begin{equation}
\begin{aligned}
    \int_X f\big(
        -[c(x, T_\theta(x)) -v(T_\theta(x))]
    \big) d\bbP
    +\int_Y f(-v(y)) d\bbQ. 
    \notag
\end{aligned}
\end{equation}
We can leverage a min-max training diagram \cite{goodfellow2014generative} for training the $T_\theta$ for each $v$, since there is no closed-form for the optimal $T$. By parameterizing the $v$ as a neural potential network $v_\varphi$, the networks' objective $\mathcal L_{T,v}$ is,
\begin{equation}
\begin{aligned}
\inf_{v_{\varphi}} \bigg[ \int_X f\Big(\sup_{T_\theta} \big\{ c(x, T_\theta(x)) - v_{\varphi}(T_\theta(x)) \big\} \Big) d\mathbb{P}\\
+ \int_Y f(-v_{\varphi}(y)) d\mathbb{Q} \bigg].
\label{eq: uot-objective}
\end{aligned}
\end{equation}

We discuss differences with GANs \cite{goodfellow2014generative} in Suppl. Sect. {\color{iccvblue} 3}.
\subsubsection{Pansharpening Regulized UOT Cost Function}
\label{sect: pansharpening-uot-cost}
However, the relaxed marginals do not admit the pansharpening regularizations, that is, the sampled (or fused) image may not meet the spatial or spectral consistency from PAN and LRMS. Hence, we introduce a novel regularization term and embed it into the UOT cost function. Following \cite{wu2021voplus,cao2024zero}, the HRMS $y$ can be degraded back to PAN $p$ and LRMS $m$ in spatial and spectral aspects, respectively:
\begin{equation}
\begin{cases}
    \text{Spatial}: &\tilde p=yBS\\
    \text{Spectral }: &\tilde m=y-yB\odot(\hat p\oslash \hat p_L),
\end{cases}
\label{eq: degraded-pan-lrms}
\end{equation}
where $B\in \mathbb R^{HW\times HW}$ is matrixization blurring operator, $S \in \mathbb R^{HW\times hw}$ denotes decimation matrix consisting of sparse components, $\hat p$ can be regarded as a function of LRMS bands and PAN and pre-processed by spectral matching \cite{unger2007introductory}, and $\hat p_L$ is obtained by exploiting MTF-filters \cite{wu2021voplus}. $\odot$ and $\oslash$ are elementwise product and division. Based on this, we can modify the original UOT cost in Eq.~\eqref{eq: uot-objective} into \textit{pansharpening-regularized cost},
\begin{equation}
    \tilde c(x,y)=c(x,y)+\|p-\tilde p\|_2^2 + \|m-\tilde m\|_2^2,
    \label{eq: pansharpening-ot-cost}
\end{equation}
where $\tilde p$ and $\tilde m$ are obtained from Eq.~\eqref{eq: degraded-pan-lrms}.

\begin{proposition}[Saddle points of pansharpening-regularized UOT provide the OT maps] \label{prop: saddle-points-give-ot-map}
    For any optimal potential function $v^* \in \mathop{\arg\sup_{v}} \mathcal L_{T,v}$ (\textit{i.e.}, Eq.~\eqref{eq: uot-objective}), it provides the OT map $T^*$ which holds
    \begin{equation}
        T^*\in \mathop{\arg\min}_T \mathcal L (T, v^*).
    \end{equation}
\end{proposition}
The proof is provided in Suppl. Sect. {\color{iccvblue} 2.2}. Prop.~\ref{prop: saddle-points-give-ot-map} gives that the although pansharpening task introduces additional cost in the original OUT cost, there is an optimal pair that can be acquired by solving the min-max problem in Eq.~\eqref{eq: uot-objective}. By using the optimal pair, an OT map from $\bbP$ to $\bbQ$ can be constituted.

\subsection{Taming FM with UOT into One-step Generator}

Recall that our initial goal is to turn the multistep flow matching into a one-step generator. In this subsection, we proposed a novel algorithm that embeds the UOT min-max problem into flow matching learning, thereby forming the \textit{simulation-free} training and \textit{one-step} sampling framework.

Note that in original flow matching, we need to learn a velocity network $s_{\theta,t}$ to approximate the true velocity $y-x$, that is, $y_0:=x \sim \mathcal N(0, \mathbf I)$ and $y_1:=y \sim \bbQ_{HRMS}$). Note that we \textit{change the start point of FM $y_0$ into LRMS (not Gaussian)} since it eases the UOT cost learning and sets an appropriate estimation to sample \cite{cao2024neural}. In flow matching, if the velocity $s_t$ is obtained, we can access the predicted $y_1 = s_t + y_0$. Thus, we can convert the velocity network into a one-step generator: $T_{\theta,t}:=s_{\theta,t} +y_0$, where $T$ is the UOT mapping network.
Note that the flow condition is unchanged, \textit{i.e.}, PAN, and LRMS. Naturally, the flow learning objective is converted into $y_1$-predicted under the UOT context. Note that we keep the flow matching loss~\eqref{eq: flow-matching-loss}, since flow loss constrains the flow path $\{y_t\}_{t=0}^1$ into a near-straight path \cite{liu2022flow}, forming a \textit{simulation-free} training and further stabilizes the UOT training.
One question is whether converting the network's objective may affect the network's performance. As shown in EDM \cite{karras2022elucidating}, re-weighting the output of the network into $y_1$ has proven to be effective in learning a better diffusion model. For the potential network, due to the timestep introduced into OTFM, we condition the potential network \textit{w.r.t} to $t$, that is $v_{\varphi,t}:=v_\varphi$.

Finally, we can summarize the training algorithm in Algo.~\ref{algo: train-otfm}. After the mapping network is well-trained, we can use the mapping network to directly sample in \textit{one step}.

\definecolor{green1}{RGB}{158, 200, 185}
\newcommand{\up}{$\uparrow$}
\newcommand{\down}{$\downarrow$}

\SetCommentSty{myCommentStyle}
\newcommand\myCommentStyle[1]{\textcolor{green1}{#1}}
\SetKwComment{Comment}{$\triangleright$ }{}
\SetKwComment{tcp}{$\triangleright$ }{}
\newcommand{\linecomment}[1]{\tcp*[f]{#1}}
\normalem

\begin{algorithm}[!t]
\caption{Training algorithm of OTFM.}
\label{algo: train-otfm}
\KwIn{The coupled HRMS, PAN, and LRMS distribution $\bbQ \times \mathbb C$, convex function $f$, one-step flow mapping network $T_\theta(\cdot)$, potential network $v_\varphi$.}
\KwOut{Trained flow mapping network $T_{\theta^*}(\cdot)$}

\While{Not Converged}{
    $Y_1, P, M\gets \bbQ \times \mathbb C$. \linecomment{A batch of training data}\\
    $Y_0\gets M$ \linecomment{$y_0$ is LRMS, not a Gaussian}\\
    $t\gets \mathcal U(0,1)$, $Y_t\gets tY_0+(1-t)Y_1$. \linecomment{Flow state}\\
    \Comment{Flow matching loss}
    $\mathcal L_{flow}$=$
    \frac{1}{|Y_0|}\sum_{y_0\in Y_0, y_1\in Y_1}\|s_{\theta,t}(y_t,t,m,p)-(y_1-y_0)\|_2^2$.\\
    $\hat Y_1=s_{\theta,t}+Y_t.$ \linecomment{Get OT mapped $y$}\\
    \Comment{UOT mapping network loss, $\tilde c$ is in Eq.~\eqref{eq: pansharpening-ot-cost}}
    $\mathcal L_{T}=\frac {1}{|Y_0|}\sum_{y_0,y_1,\hat y_1}[\tilde c(y_0,\hat y_1) - v_{\varphi,t}(\hat y_1,t)].$\\
    Update $\theta$ using loss $\mathcal L_{flow} + \mathcal L_{T}$.\\
    \Comment{Potential network loss}
    $\mathcal L_v=\frac{1}{|Y_0|}\sum_{y_0,
    \hat y_1}[f(-\tilde c(y_0,\hat y_1)+v_{\varphi,t}(\hat y_1,t)] + \frac{1}{|Y_1|}\sum_{y_1}[f(-v_{\varphi,t}(y_1,t)]$.\\
    Update $\varphi$ using loss $\mathcal L_v$.\\
}
$T_{\theta^*}\gets T_\theta$. \linecomment{Trained mapping network $T_{\theta^*}$}\\
\end{algorithm}

\subsection{Deep Network for OTFM}
Network designing is a vital part of OTFM, to suit pansharpening task that provides two images as conditions, we design an effecient network architecture for the mapping network $T_\theta$ and the potential network $v_\varphi$. 

As shown in Fig.~\ref{fig:otfm-framework} (b), since the U-net \cite{ronneberger2015u} is widely used in diffusion methods, we adopt a UNet-like structure for the mapping network. Specifically, the network consists of three encoder layers and decoder layers. Each encoder layer is followed by a downsampling operation with a factor of 2, and each decoder layer is followed by an upsampling operation to adjust spatial resolution and channel quantity during feature transmission.
The encoder and decoder are each composed of two stacked basic blocks. 
Each basic block adopts attention and FFN blocks. For an efficient attention computation, we leverage the neighborhood attention \cite{hassani2022dilated} to obtain the global feature response. As for the FFN, it is implemented as three MLPs. 
Note that, the mapping network is additionally conditioned on PAN and LRMS. To fulfill the condition injection, we utilize the DiT \cite{peebles2023scalable} AdaLN-zero conditioning,
\begin{equation}
\begin{cases}
    \alpha,\beta,\gamma=MLP(LN([m,p])),\\
    x_{op}=(1+\alpha)\cdot op(x)  + \beta, \\
    x_{out}=x+\gamma \cdot x_{op},
\end{cases}
\end{equation}
where $x$ is the feature, $op\in \{\text{attention}, \text{FFN}\}$, $[\cdot]$ is the channel concatenation, and $LN$ denotes the layer-norm.
The encoder layer encodes input features, downsamples them, and forwards the output to the corresponding decoder layer through skip connections. The decoder layer decodes features, integrates them with skip-connected features, and upsamples the result.
Convolutions are employed at the network's front and back to ensure feature dimensions align with the network or target image's channel specifications.

To leverage the potential network, we adopt a patch GAN discriminator \cite{isola2017image} into a $t$-conditioned network. Specifically, as illustrated in Fig.~\ref{fig:otfm-framework} (a), the potential network comprises three convolutional blocks, each consisting of a ``Conv+BN+LReLU'' layer sequence. The timestep $t$ is encoded using a cosine function and integrated with the feature map. Finally, the resulting feature is flattened and averaged to yield the potential value $v_{\varphi}$.

\begin{figure*}[t]
    \centering
    \includegraphics[width=\linewidth]{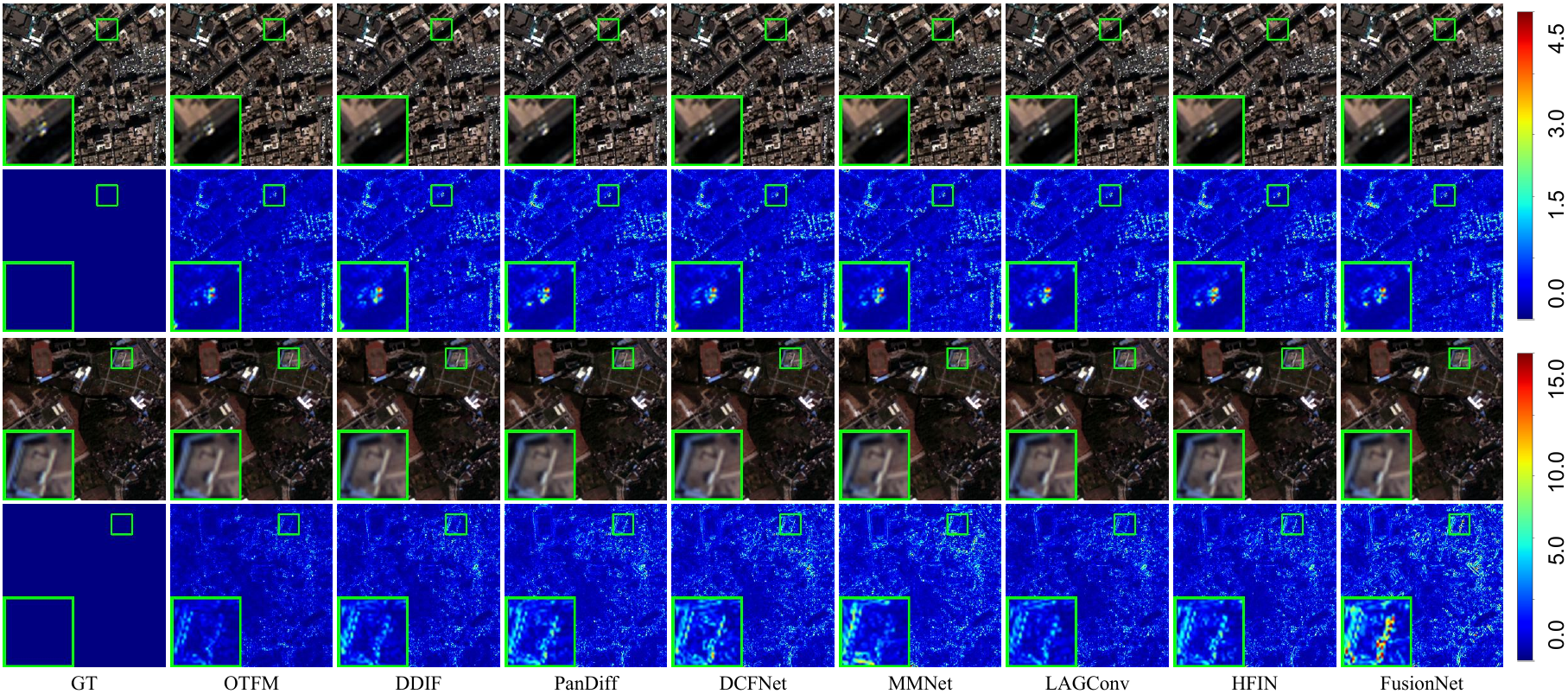}
    \caption{Visual comparisons on WV3 (1-2 rows) and GaoFen-2 (3-4 rows) cases. The second the fourth rows are error maps.}
    \label{fig: reduced_wv3}
    \vspace{-6pt}
\end{figure*}

\section{Experiments}

In this section, we will provide a comprehensive overview of the experimentation settings including the implementation details and datasets. After that, we will present the benchmarks employed to evaluate the performance of our approach to pansharpening. Finally, the main results and ablation studies will be provided to quantitatively illustrate the effectiveness of the proposed method.

\subsection{Implementation Details}
The mapping network and the potential network are optimized with AdamW optimizer \cite{loshchilov2017decoupled} with learning rates set to $2e^{-4}$ and $1e^{-4}$, respectively. We use exponential moving average for both models with the decay ratio set to 0.99. The maximum training steps are set to 100k, and the batch size is 52. The model was trained for one day with two NVIDIA GeForce RTX 4090 GPUs. 

\subsection{Datasets}
To better validate the effectiveness of the proposed OTFM, we conducted training and testing on a public dataset, Pancollection\footnote[1] {\url{https://liangjiandeng.github.io/PanCollection.html}.}, which includes data from three types of remote sensing satellites: WV3 (8 bands), GF2 (4 bands), and QB (4 bands). More details about the datasets are provided in Suppl. Sect. {\color{iccvblue} 4.1}.

\subsection{Benchmark and Metrics}
To evaluate the performance of our OTFM, we compare it with various state-of-the-art (SOTA) methods of pansharpening (on WV3, GF2, and QB datasets).  We choose two traditional methods: MTF-GLP-FS \cite{mtf-glp-fs}, BT-H \cite{BT-H}, as well as nine deep learning-based methods: DiCNN \cite{DICNN}, FusionNet \cite{deng2020detail}, LAGConv \cite{jin2022lagconv}, DCFNet \cite{dcfnet}, MMNet \cite{yan2022memory}, Pandiff \cite{meng2023pandiff}, HFIN \cite{HFIN}, and DDIF~\cite{cao2023ddif}. 
We utilize four common metrics to evaluate the results of these methods on the pansharpening reduced-resolution datasets, including SAM~\cite{yuhas1992discrimination}, ERGAS~\cite{wald2002data}, Q2n~\cite{garzelli2009hypercomplex}, and SCC~\cite{zhou1998wavelet}. Meanwhile, the $D_{\lambda}$, $D_s$, and HQNR~\cite{arienzo2022full} indices are used to evaluate the full-resolution datasets.

\newcommand{\best}[1]{{\cellcolor{red!10}\textbf{#1}}}
\newcommand{\second}[1]{{\cellcolor{blue!10}\textbf{#1}}}

\begin{table*}[t]
	\centering
	\caption{Quantitative results on WV3 dataset. The best results are in \sethlcolor{red!10}\hl{red} and the second best results are in \sethlcolor{blue!10}\hl{blue}.}
	\label{tab:pansharpening}
    
	\resizebox{\linewidth}{!}{
		\begin{tabular}{lcccc|ccc|c}
			\toprule
			\multicolumn{1}{l}{\multirow{2}{*}{Methods}} &
			\multicolumn{4}{c|}{Reduced-Resolution (RR): Avg$\pm$std} &
			\multicolumn{3}{c|}{Full-Resolution (FR): Avg$\pm$std}  & 
            \\
			\multicolumn{1}{l}{} &
			SAM ($\downarrow$) & 
			ERGAS ($\downarrow$) & 
			Q2n ($\uparrow$) & 
			SCC ($\uparrow$) & 
			$D_\lambda$ ($\downarrow$) & 
			$D_s$ ($\downarrow$) & 
			HQNR ($\uparrow$) & 
            NFEs
            \\ 
            \cline{2-9}
			MTF-GLP-FS~\cite{mtf-glp-fs} &
			5.32$\pm$1.65 &
			4.65$\pm$1.44 &
			0.818$\pm$0.101 &
			0.898$\pm$0.047 &
			0.021$\pm$0.008 &
			0.063$\pm$0.028 &
			0.918$\pm$0.035 &
            --
            \\
			BT-H~\cite{BT-H} &
			4.90$\pm$1.30 &
			4.52$\pm$1.33 &
			0.818$\pm$0.102 &
			0.924$\pm$0.024 &
			0.057$\pm$0.023 &
			0.081$\pm$0.037 &
			0.867$\pm$0.054 &
            --
            \\
            \hline
			DiCNN~\cite{DICNN} &
			3.59$\pm$0.76 &
			2.67$\pm$0.66 &
			0.900$\pm$0.087 &
			0.976$\pm$0.007 &
			0.036$\pm$0.011 &
			0.046$\pm$0.018 &
			0.920$\pm$0.026 &
            1 
            \\
			FusionNet~\cite{deng2020detail} &
			3.33$\pm$0.70 &
			2.47$\pm$0.64 &
			0.904$\pm$0.090 &
			0.981$\pm$0.007 &
			\second{0.024$\pm$0.009} &
			{0.036$\pm$0.014} &
			{0.941$\pm$0.020} &
            1
            \\
			LAGConv~\cite{jin2022lagconv} &
			3.10$\pm$0.56 &
			2.30$\pm$0.61 &
			0.910$\pm$0.091 &
			0.984$\pm$0.007 &
			0.037$\pm$0.015 &
			0.042$\pm$0.015 &
			0.923$\pm$0.025 &
            1
            \\
			DCFNet~\cite{dcfnet} &
			3.03$\pm$0.74 &
			{2.16$\pm$0.46} &
			0.905$\pm$0.088 &
			\second{0.986$\pm$0.004} &
			0.078$\pm$0.081 &
			0.051$\pm$0.034 &
			0.877$\pm$0.101 &
            1
            \\
            MMNet~\cite{yan2022memory} 
            & 3.08$\pm$0.64 
            & 2.34$\pm$0.63 
            & 0.916$\pm$0.086
            & 0.983$\pm$0.006
            & 0.054$\pm$0.023 
            & {0.034$\pm$0.012}
            & 0.914$\pm$0.028 
            & 1
            \\
            
            HFIN~\cite{HFIN}
                & 3.08$\pm$0.63
			& 2.31$\pm$0.58
			& 0.912$\pm$0.089
			& 0.984$\pm$0.006
			& 0.025$\pm$0.008
			& {0.043$\pm$0.017}
			& {0.934$\pm$0.024}
            & 1
            \\
            \hline
			PanDiff~\cite{meng2023pandiff}
			& 3.30$\pm$0.60
			& 2.47$\pm$0.58
			& 0.898$\pm$0.088
			& 0.980$\pm$0.006
			& 0.027$\pm$0.012
			& 0.054$\pm$0.026
			& 0.920$\pm$0.036
            & 1000
			\\
            DDIF~\cite{cao2023ddif}
            & \best{2.74$\pm$0.51}
            & \best{2.01$\pm$0.45}
            & \best{0.920$\pm$0.082}
            & \best{0.988$\pm$0.003}
            & 0.026$\pm$0.008
            & \second{0.023$\pm$0.008}
            & \second{0.952$\pm$0.017}
            & 25
            \\
            OTFM (ours) & \second{2.76$\pm$0.52}
            & \second{2.03$\pm$0.44}
            & \second{0.919$\pm$0.083}
            & \best{0.988$\pm$0.003}
            & \best{0.018$\pm$0.007}
            & \best{0.029$\pm$0.006}
            & \best{0.954$\pm$0.012}
            & 1
            \\
			\bottomrule
		\end{tabular}%
	}
    \vspace{-1em}
\end{table*}

\begin{table*}[t]
    \centering 
	\setlength{\tabcolsep}{5pt}
	\renewcommand\arraystretch{1}
    \caption{Quantitative results on the GaoFen-2 datasets. The best results are in \sethlcolor{red!10}\hl{red} and the second best results are in \sethlcolor{blue!10}\hl{blue}.}
    \resizebox{\linewidth}{!}{
    \begin{tabular}{lcccc|ccc|c}
        \toprule
        \multirow{2}{*}{Method}& \multicolumn{4}{c|}{Reduced-Resolution (RR): Avg$\pm$std} 
        & \multicolumn{3}{c}{Full-Resolution (FR): Avg$\pm$std} 
        \\  
			& SAM ($\downarrow$) &
			ERGAS ($\downarrow$) &
			Q2n ($\uparrow$) &
			SCC ($\uparrow$) 
                &
			$D_\lambda$ ($\downarrow$) &
			$D_s$ ($\downarrow$) &
			HQNR ($\uparrow$)
                & NFEs
                \\
        \cline{2-9}
        
        MTF-GLP-FS~\cite{mtf-glp-fs} &
			1.68$\pm$0.35 &
			1.60$\pm$0.35 &
			0.891$\pm$0.026 &
			0.939$\pm$0.020 
                & 0.035$\pm$0.014 &
			0.143$\pm$0.028 &
			0.823$\pm$0.035 &
                --
            \\
			BT-H~\cite{BT-H} &
			1.68$\pm$0.32 &
			1.55$\pm$0.36 &
			0.909$\pm$0.029 &
			0.951$\pm$0.015 
                & 0.060$\pm$0.025 &
			0.131$\pm$0.019 &
			0.817$\pm$0.031 &
                --
            \\
            \hline

			DiCNN~\cite{DICNN} &
			1.05$\pm$0.23 &
			1.08$\pm$0.25 &
			0.959$\pm$0.010 &
			0.977$\pm$0.006 
                & 0.041$\pm$0.012 &
			0.099$\pm$0.013 &
			0.864$\pm$0.017 &
                1
            \\
			FusionNet~\cite{deng2020detail} &
			0.97$\pm$0.21 &
			0.99$\pm$0.22 &
			0.964$\pm$0.009 &
			0.981$\pm$0.005 
                &0.040$\pm$0.013 &
			0.101$\pm$0.013 &
			0.863$\pm$0.018 &
                1
            \\
			LAGConv~\cite{jin2022lagconv} &
			\second{0.78$\pm$0.15} &
			0.69$\pm$0.11 &
			0.980$\pm$0.009 &
			\second{0.991$\pm$0.002} 
            & 0.032$\pm$0.013 &
			0.079$\pm$0.014 &
			0.891$\pm$0.020 &
                1
            \\
			DCFNet~\cite{dcfnet} &
			0.89$\pm$0.16 &
			0.81$\pm$0.14 &
			0.973$\pm$0.010 &
			0.985$\pm$0.002 
                &0.023$\pm$0.012 &
			{0.066$\pm$0.010} &
			{0.912$\pm$0.012} &
                1
            \\
            MMNet~\cite{yan2022memory}
            & 0.99$\pm$0.14
            & 0.81$\pm$0.12 
            & 0.969$\pm$0.020
            & 0.986$\pm$0.002 
            & 0.043$\pm$0.030 
            & 0.103$\pm$0.013
            & 0.858$\pm$0.027  &
            1
            \\
                HFIN~\cite{HFIN}
			& {0.84$\pm$0.15}
			& 0.73$\pm$0.13
			& 0.977$\pm$0.011
			& 0.990$\pm$0.002
                & 0.027$\pm$0.020
			& 0.062$\pm$0.010
			& 0.912$\pm$0.018
                & 1
                \\

                \hline
                
                PanDiff~\cite{meng2023pandiff}
			& 0.89$\pm$0.12
			& 0.75$\pm$0.10
			& 0.979$\pm$0.010
			& 0.989$\pm$0.002
			& 0.027$\pm$0.020
			& 0.073$\pm$0.010
			& 0.903$\pm$0.021 
                & 1000
            \\

                DDIF~\cite{cao2023ddif}
                & \best{0.64$\pm$0.12}
                & \best{0.57$\pm$0.10}
                & \second{0.985$\pm$0.008} 
                & 0.986$\pm$0.003 
                & \second{0.020$\pm$0.011} 
                & \best{0.041$\pm$0.010}
                & \second{0.939$\pm$0.014}
                & 25
            \\
			Proposed &
			\best{0.64$\pm$0.12} &
			\second{0.57$\pm$0.11} &
			\best{0.985$\pm$0.007} &
			\best{0.993$\pm$0.001} &
			\best{0.019$\pm$0.009} &
			\second{0.042$\pm$0.008} &
			\best{0.939$\pm$0.010} &
                1
            \\
        \bottomrule
    \end{tabular}}
    \vspace{-1em}
    \label{tab: gf2_reduced}
\end{table*}

\begin{table}[t]
    \centering 
	\setlength{\tabcolsep}{1pt}
    \caption{Results on the QuickBird reduced-resolution datasets. The best results are in \sethlcolor{red!10}\hl{red} and the second best results are in \sethlcolor{blue!10}\hl{blue}.}
    \resizebox{\linewidth}{!}{
    \begin{tabular}{lcccc}
        \toprule
        \multirow{2}{*}{Method}& \multicolumn{4}{c}{Reduced-Resolution (RR): Avg$\pm$std} 
        \\  
        
			& SAM ($\downarrow$) & 
			ERGAS ($\downarrow$) &
			Q2n ($\uparrow$) & 
			SCC ($\uparrow$) 
                \\
        \midrule
        
        MTF-GLP-FS~\cite{mtf-glp-fs}      & 8.11$\pm$1.95 & 7.51$\pm$0.79 & 0.829$\pm$0.090 & 0.899$\pm$0.019 
        \\ 
        BT-H~\cite{BT-H}            & 7.19$\pm$1.55 & 7.40$\pm$0.84 & 0.833$\pm$0.088 & 0.916$\pm$0.015  
        \\ 
        \hline
        DiCNN~\cite{DICNN}           & 5.38$\pm$1.03 & 5.14$\pm$0.49 & 0.904$\pm$0.094 & 0.962$\pm$0.013 
        \\
        FusionNet~\cite{deng2020detail}       & 4.92$\pm$0.91 & 4.16$\pm$0.32 & 0.925$\pm$0.090 & 0.976$\pm$0.010 
        \\
        LAGConv~\cite{jin2022lagconv}          & 4.55$\pm$0.83 & 3.83$\pm$0.42 & 0.933$\pm$0.088 & 0.981$\pm$0.009 
        \\
        DCFNet~\cite{dcfnet}          & {4.54$\pm$0.74} & 3.83$\pm$0.29 & 0.933$\pm$0.090 & 0.974$\pm$0.010 
        \\ 
        MMNet~\cite{yan2022memory} & 4.56$\pm$0.73 & {3.67$\pm$0.30} & 0.934$\pm$0.094 & 0.983$\pm$0.007 
        \\ 
        HFIN~\cite{HFIN} & 4.54$\pm$0.81 & 3.81$\pm$0.32 & {0.934$\pm$0.085} &  0.980$\pm$0.010 
        \\ 
        \midrule
        PanDiff~\cite{meng2023pandiff} & 4.58$\pm$0.74 & 3.74$\pm$0.31 & 0.935$\pm$0.090 &  0.982$\pm$0.090 
        \\
        DDIF~\cite{cao2023ddif} & \best{4.35$\pm$0.73} & \best{3.52$\pm$0.27} & \second{0.938$\pm$0.090} & \second{0.984$\pm$0.079} \\
        Proposed  & \second{4.39$\pm$0.76} & \second{3.54$\pm$0.28} & \best{0.938$\pm$0.088} & \best{0.985$\pm$0.070} 
        \\ 
        \bottomrule
    \end{tabular}
    }
    \label{tab: qb_reduced}
\end{table}

\begin{table}[t]
    \centering
	\setlength{\tabcolsep}{2pt}
    \caption{Ablation studies on different one-step generators, the proposed pansharpening-regularization, and comparisons with previous OT and GAN methods.}
    \label{tab: ablation}
    \resizebox{\linewidth}{!}{
    \begin{tabular}{l|cccc|c}
    \toprule
        Methods & SAM ($\downarrow$) & ERGAS ($\downarrow$) & Q2n ($\uparrow$) & SCC ($\uparrow$) & HQNR ($\uparrow$) \\
        \hline
        (a) Reflow \cite{liu2022flow} & 3.18 & 2.32 & 0.915 & 0.983 & 0.920 \\
        (b) CM \cite{song2023consistency} & 2.88 & 2.12 & 0.916 & 0.985 & 0.936 \\
        \hline 
        (c) w/o UOT (FM) & 2.83 & 2.10 & 0.917 & 0.986 & 0.941\\
        (d) w/o UOT (Diff) & 2.87 & 2.12 & 0.915 & 0.986 & 0.942 \\
        (e) w/o Pan-Reg cost & 2.78 & 2.08 & 0.918 & 0.986 & 0.940 \\
        \hline
        (f) FM+NOT \cite{korotin2022neural} & 2.89 & 2.11 & 0.915 & 0.985 & 0.934 \\
        (g) FM+GAN \cite{xu2024ufogen} & 2.85 & 2.12 & 0.916 & 0.986 & 0.938 \\
    \bottomrule
    \end{tabular}}
\end{table}

\begin{table}[t]
    \centering 
    \caption{Efficiency results on the $256\times 256$ resolution image. }
    \resizebox{\linewidth}{!}{
    \begin{tabular}{c|ccccc}
        \toprule
        Method & Proposed & DDIF & PanDiff & DCFNet & MMNet \\
        \hline
       latency (s) & 0.029 & 2.602 & 261.410 & 0.257 & 0.348 \\
        \bottomrule
    \end{tabular}}
    \label{tab: latency}
    \vspace{-1em}
\end{table}

\subsection{Main Results}
\textbf{Results on WV3.} 
We evaluated the performance of OTFM on 20 test images from both the reduced-resolution dataset and the full-resolution dataset of WV3, with the results presented in Tab.~\ref{tab:pansharpening}. Compared to two traditional methods and eight SOTA DL-based methods, the proposed OTFM performs the best overall performance on the reduced-resolution dataset. To illustrate the performance differences among each method clearly, we present the fusion result images and the error maps for some methods in Fig.~\ref{fig: reduced_wv3} with a specific location zoomed in. Additionally, our OTFM achieves SOTA on the real full-resolution dataset, achieving an HQNR of 0.954. With only one sampling step, OTFM can generate high-quality HRMS better than DDIF.

\textbf{Results on GF2.}
Tab.~\ref{tab: gf2_reduced} presents the test results for the reduced-resolution 20 test images GF2 dataset. Our OTFM achieves SOTA performance, and as shown in Fig.~\ref{fig: reduced_wv3}, the fusion results from OTFM exhibit reduced spectral distortion while retaining more spatial details compared to other methods, which is comparable with 25-sampling-step DDIF.
Furthermore, OTFM outperforms previous methods on the real full-resolution dataset, with visual comparisons available in the Suppl. Sect. {\color{iccvblue}4.2}. This demonstrates that the proposed OTFM possesses excellent fusion performance, as well as effectiveness on real datasets.

\textbf{Results on QB.}
We also conduct experiments on the QB dataset to evaluate performance at reduced resolutions. Similarly, Tab.~\ref{tab: qb_reduced} reports the quality metrics obtained from 20 randomly selected test samples from the QB dataset. Our OTFM method achieves the best overall performance across the four metrics. Additional results analysis and visual comparisons for QB dataset can be found in Suppl. Sect. {\color{iccvblue}4.3}.

\section{Ablation Study}
In this section, we ablate the proposed UOT training loss, pansharpening-regularized OT cost, and compare it with previous OT and GAN methods on the WV3 dataset.

\subsection{Compared with Reflow and Consistency Model}
Recently, Reflow \cite{liu2022flow} has been proposed to distill a multi-step sampling FM into a one-step model, while the Consistency Model (CM) \cite{song2023consistency} aims to train a model supporting one-step sampling by leveraging the ODE endpoint consistency law. For Reflow, we adhere to its distillation algorithm: first, we train an FM until convergence, then sample (noise, HRMS) pairs, and finally perform distillation on these pairs. For CM, we employ the Euler discretization approach and directly train the one-step CM. Their performances are presented in Tab.~\ref{tab: ablation} (a-b). Although both methods perform sampling in a single step, their results still fall behind. Particularly for Reflow, due to the accumulation of distillation errors, it exhibits the worst performance.

\subsection{UOT makes a One-step Generator}
We claim that UOT enables a one-step generator, which is demonstrated in this subsection by ablating the loss function associated with the proposed UOT loss. Specifically, we remove Lines 6-10 in Algo.~\ref{algo: train-otfm}, effectively training a flow matching model directly. The results are presented in Tab.~\ref{tab: ablation} (c). This model requires 100 ODE sampling steps to achieve the same pansharpening performance as our OTFM. Notably, replacing flow matching with diffusion falls back to the setting of PanDiff. We also provide the performance of diffusion within our designed deep architecture (see Tab.~\ref{tab: ablation} (d)); this variant still necessitates more sampling steps and exhibits inferior performance.

\subsection{Pansharpening-regularized Cost Enhances Generalization}
In Sect.~\ref{sect: pansharpening-uot-cost}, we introduce a task-level regularization on the UOT cost for the pansharpening task. To validate the effectiveness of this pansharpening-regularized cost, we remove the regularization term and retrain the OTFM. The results are presented in Tab.~\ref{tab: ablation} (e).
It can be seen that, with the regularization term, OTFM can reach higher reduced and full resolution metrics.
This regularization brings full-resolution HQNR from 0.940 to 0.954.
Thus, we can conclude that the proposed pansharpening OT regularization enhances the model's generalization capability.

\subsection{UOT beats previous OT and GAN}
Recent OT approaches, such as NOT \cite{korotin2022neural} and the GAN-based method \cite{xu2024ufogen}, have demonstrated strong performance in image translation tasks. We integrate these methods with FM to verify the effectiveness of the proposed OTFM. Both approaches follow a one-step generation process, and their results are presented in Tab.~\ref{tab: ablation} (f-g). 
It can be observed that our OTFM outperforms them on both reduced and full resolution metrics. Additionally, we found that the training loss of GAN methods is less stable compared to UOT. We conjecture that this may be due to the GAN discriminative instability. Although the NOT+FM approach exhibits stable training, its performance is inferior.

\subsection{Inference Latency}
Inference latency is a vital aspect to examine the efficiency of pansharpening methods.
In Tab.~\ref{tab: latency}, we include the comparisons with previous SOTA multi-step diffusion methods: DDIF and PanDiff, as well as previous regression-based methods: DCFNet and MMNet. By testing them and ours under $256\times 256$ resolution, OTFM can fuse the HRMS using only 0.029 seconds due to the efficient one-step sampling and designs, much faster than previous diffusion-based methods.

\section{Conclusion}
This work introduces unbalanced optimal transport (UOT) into flow matching to address multi-step sampling limitations. By integrating pansharpening-specific constraints into UOT’s cost function, we mitigate restrictive marginal conditions and enable one-step generation without teacher-student networks or extensive (noise, HRMS) pair sampling. Our method matches the performance of multi-step diffusion approaches on pansharpening datasets while achieving $\sim 10\times$ faster inference.

{\small
\bibliographystyle{ieee_fullname}
\bibliography{egbib}
}

\end{document}